\documentclass[letterpaper, 10 pt, conference]{ieeeconf}  

\IEEEoverridecommandlockouts                              
\usepackage{amsmath}
\usepackage{amssymb}
\usepackage{amsthm}
\usepackage{bm}
\usepackage{color}
\usepackage{url}
\usepackage[hidelinks]{hyperref}
\usepackage{lipsum}

\usepackage[boxed,noend]{algorithm2e}
\usepackage{mathtools}

\newcommand{\E}{\mathbb{E}}
\renewcommand{\Pr}{\mathbb{P}}
\newcommand{\cI}{\mathcal{I}}
\newcommand{\cS}{\mathcal{S}}
\newcommand{\cA}{\mathcal{A}}
\newcommand{\cM}{\mathcal{M}}
\newcommand{\cO}{\mathcal{O}}
\newcommand{\cV}{\mathcal{V}}
\newcommand{\bcA}{\Bar{\cA}}

\newcommand{\blue}[1]{{#1}} 

\theoremstyle{definition}

\theoremstyle{remark}
\newtheorem{remark}{Remark}

\theoremstyle{plain}

\allowdisplaybreaks
\mathtoolsset{showonlyrefs}

\DeclareMathOperator*{\argmax}{arg\,max}

\title{\LARGE \bf
A Novel Point-based Algorithm for Multi-agent Control Using the Common Information Approach
}

\author{Dengwang Tang, Ashutosh Nayyar, Rahul Jain
\thanks{\blue{Shorter version submitted to the 62nd IEEE Conference on Decision and Control (CDC2023).}}
\thanks{This work is supported by ONR award N00014-20-1-2258 and NSF awards ECCS-2025732 and ECCS-1750041.}
\thanks{D. Tang, A. Nayyar, and R. Jain are with Ming Hsieh Department of Electrical and Computer Engineering,
        University of Southern California,
        Los Angeles, CA 90089-2560. Email:
        {\tt\small \{dengwang, ashutosh.nayyar, rahul.jain\}@usc.edu}}%
}

\begin{document}

\maketitle

\thispagestyle{plain}  
\pagestyle{plain}  

\begin{abstract}
The Common Information (CI) approach provides a systematic way to transform a multi-agent stochastic control problem to a single-agent partially observed Markov decision problem (POMDP) called the coordinator's POMDP. However, such a POMDP can be hard to solve due to its extraordinarily large action space. We propose a new algorithm for multi-agent stochastic control problems, called coordinator's heuristic search value iteration (CHSVI), that combines the CI approach and point-based POMDP algorithms for large action spaces. We demonstrate the algorithm through optimally solving several benchmark problems.
\end{abstract}

\section{Introduction}
Multi-agent control problems arise in a number of applications including in  teams of autonomous agents or robots carrying out a mission, in communication networks with multiple  users  and in distributed systems like the smart grid. The problem typically involves multiple agents with potentially different information taking actions and interacting with a dynamic system. In cooperative multi-agent problems, the agents' goal is to optimize a shared performance metric. 

\blue{
In this work, we assume that a control strategy can be determined offline collectively before interacting with the system (or \emph{centralized planning and decentralized execution} as in Dec-POMDP literature). 
Even with centralized planning, multi-agent control problems can be significantly more difficult than their single-agent counterpart due to a variety of reasons: (1) due to the interdependent nature of all agents' information, actions, and the system evolution dynamics, we need to determine the strategies of all agents at the same time; (2) the decision making process of an agent involves not only the estimation of the underlying state but also the estimation of other agent's information, since it is crucial to predict other agents' actions. By the same token, an agent also needs to understand what other agents may believe of the information and action of herself and so on, resulting in the need to form a complicated hierarchy of beliefs.
} 

There are several structural approaches to transform or decompose a multi-agent control problem into single-agent control problems. One of these approaches is the common information (CI) approach \cite{nayyar2013decentralized}. In this approach, a multi-agent control problem is transformed into a single-agent partially observed Markov decision problem (POMDP) by assuming the presence of a fictitious player, called \emph{the coordinator}. 
At each time, the information available for each agent is partitioned into two parts, the \emph{common information} and the \emph{private information}. 
At each time, instead of letting each individual agent decide on their actions, the coordinator selects a \emph{prescription} for each agent, which is a  mapping from that agent's private information to its actions. The choice of prescription is based solely on the common information. In principle, the coordinator's problem can be solved using single-agent POMDP algorithms.

Single-agent POMDPs are nevertheless not easy to solve. Exact value iteration methods of solving POMDPs through updating a set of support vectors of the value function, namely \emph{$\alpha$-vectors}, were introduced in \cite{sondik1978optimal,cassandra1997incremental}. However, these methods were practical only for problems with very few states \cite{shani2013survey}. A class of approximate solution methods, called \emph{point-based} methods \cite{pbvi,hsvi,sarsop} have been quite successful in approximately solving POMDP problems with hundreds or thousands of states \blue{(e.g. PBVI \cite{pbvi} is the first algorithm to demonstrate great performance on the $870$-state \emph{Tag} problem. HSVI2 \cite{hsvi2} is capable of tackling the $12,545$-state \emph{RockSample}[7,8] problem). }
These methods are built on the following premise: Instead of computing all $\alpha$-vectors, one can efficiently approximate the Bellman update through \emph{point-based backup} procedures, i.e., computing a relatively small subset of $\alpha$-vectors representing the gradients of the updated value function at certain belief points (see \cite{shani2013survey} for a  comprehensive survey and a tutorial on such methods).

However, compared to a typical single-agent POMDP, a coordinator's POMDP  has an \emph{astronomically} large action space, since its actions are mappings. (Even a seemingly small problem like \texttt{DecTiger}~(2,1,$\beta$) (see Appendix \ref{dectiger}) has millions of actions.) The huge action space creates major challenges in the use of state-of-the-art point-based methods to coordinator's POMDPs because: (1) The backup procedure requires solving discrete optimization problems over the action space. (2) The belief exploration processes of certain point-based methods \cite{hsvi,sarsop} also require solving discrete optimization problems over the action space. Therefore, even though the CI approach transforms a multi-agent problem into a single-agent one, there's still a need for a specialized algorithm to solve these specialized single-agent problems.

In this work, we present a new algorithm for multi-agent control that combines the CI approach with point-based POMDP solution methods for large action spaces. We demonstrate the performance of the new algorithm on several benchmark problems with huge action spaces.

\textbf{Related Work:} \blue{The CI approach \cite{nayyar2013decentralized} has been applied in many settings. In \cite{nayyar2013optimal,mahajan2013optimal}, the authors applied the approach to establish structural results for certain classes of multi-agent communication problems. In \cite{hamidsuffinfo}, the authors extended the approach to include the compression of private information. In \cite{subramanian2022approximate,zhang2019online,kao2022common}, the authors proposed the idea of combining the CI approach with approximate planning and learning methods.}
While many theoretical results have been developed using the CI approach \cite{nayyar2013decentralized,mahajan2013optimal,hamidsuffinfo,kao2022common}, there is a lack of efficient planning algorithms based on this approach. As a result, empirical results have been established either in the case of very small private information spaces \cite{zhang2019online}, or through machine learning techniques \cite{foerster2019bayesian}.

The model of Dec-POMDP is a multi-agent extension of POMDP that has been studied extensively (see \cite{amato2013decentralized} for a survey), where point-based methods have been applied (e.g. \cite{kumar2010point,macdermed2013point,dibangoye2016optimally}). In many Dec-POMDP algorithms, the policy are represented as policy trees \cite{dibangoye2016optimally}, whose size grows exponentially in time horizon. Therefore, such algorithms require a huge amount of memory and can only solve small finite horizon problems. There have also been Dec-POMDP algorithms designed for infinite horizon problems \cite{pajarinen2011periodic,dibangoye2014error}. However, those algorithms do not provide an optimality guarantee. 

Point-based methods have been applied to multi-agent problems with partial history sharing. For example, in \cite{adhikari2021point}, the authors applied point-based methods to finite horizon problem with communication between agents. However, to the best of our knowledge, our work is the first to combine the CI approach with point-based POMDP methods on general infinite-horizon multi-agent problems. 

\textbf{Contributions:} (1) In the context of works related to the CI approach, we present the first practical algorithm to solve a general coordinator's POMDP with large action spaces. (2) In the context of the Dec-POMDP literature, we provide a memory-efficient anytime algorithm for infinite horizon problems with an upper and lower bound for the value of the output policy.

\section{Preliminaries}
\subsection{Problem Formulation}

\subsubsection{Multi-agent Control with Partial History Sharing}\label{sec:modelmulti}
We consider an infinite-horizon multi-agent control model characterized by a tuple $\mathcal{E}= (\cI, \cS, \cA,  \cM, \cO, b_0, \Pr, r, \beta)$ where $\cI$ is a finite set of agents;
\begin{itemize}
    \item $\cS$ is a finite set representing the state space;
    \item $\cA = \prod_{i\in\cI} \cA^i$, where $\cA^i$ is a finite set representing the action space of agent $i$;
    \item $\cM = \prod_{i\in\cI} \cM^i$, where $\cM^i$ is a finite set representing the domain of private information for agent $i$;
    \item $\cO$ is a finite set representing the domain of common observation \blue{(A common observation at time $t$ is allowed to be already observed in private by some agents before time $t$); }
    \item $b_0\in\Delta(\cS\times \cM)$ is the initial joint distribution on the state and private information of all agents; 
    \item $\Pr: \cS\times \cM\times \cA\mapsto \Delta(\cS\times \cM\times \cO)$ is the state-information joint transition kernel; 
    \item $r:\cS\times\cA\mapsto \mathbb{R}$ is the instantaneous reward;
    \item $\beta\in (0, 1)$ is the discount factor.
\end{itemize}

Player $i$'s information space at time $t$ is $\mathcal{H}_t^i = (\cO)^{t}\times\cM^i$ which represents the common observations from time $1$ to $t$ along with the private information at time $t$. Actions may or may not be commonly observed by all agents. If any $a_t^i$ is commonly observed by all agents, then it is part of the common observation $o_{t+1}$. The goal in this problem is to choose a joint strategy $\pi=(\pi^i)_{i\in\cI}, \pi^i = (\pi_t^i)_{t=0}^\infty, \pi_t^i:\mathcal{H}_t^i \mapsto \cA^i$ to maximize the total discounted-reward $J(\pi)$, i.e. 
$$\max_{\pi} \E^{\pi}\left[\sum_{t=0}^\infty \beta^t r(s_t, a_t)\right].$$ 

As shown in \cite{nayyar2013decentralized}, this model can be used to model many decentralized decision and control problems with partial history sharing  i.e. problems where certain subsets of agents' action and observation history are commonly known. 
\blue{
\begin{remark}\label{remark:suffinfo}
    In \cite{hamidsuffinfo}, the authors introduced the concept of \emph{sufficient private information}, which is a compression of private information that is sufficient for decision-making purpose. One can replace the space of private information $\cM$ in this model by the space of sufficient private information and the proposed algorithm of this work will still apply.
\end{remark}
} 

\subsubsection{Transformation into Coordinator's POMDP}\label{sec:coordpomdp}

Following the CI approach introduced in \cite{nayyar2013decentralized}, we transform the problem $\mathcal{E}$ into an equivalent POMDP problem, called the coordinator's POMDP, which can be characterized by a tuple $\Bar{\mathcal{E}} = (\Bar{\cS}, \Bar{\cA}, \cO, b_0, \Bar{\Pr}, \Bar{r}, \beta )$, where $\cO, b_0, \beta$ are the same as in the original model; $\Bar{\cS} = \cS\times \cM$ is the (augmented) state space; $\Bar{\cA} = \prod_{i\in\cI}\Bar{\cA}^i$ is the (new) action space, where $\Bar{\cA}^i$ 
is the set of mappings (called prescriptions) from $\cM^i$ to $\cA^i$; $\Bar{\Pr}: \Bar{\cS}\times \Bar{\cA}\mapsto \Delta(\Bar{\cS}\times \cO)$ is the combined state transition and observation kernel; $\Bar{r}:\Bar{\cS}\times\Bar{\cA}\mapsto\mathbb{R}$ is the instantaneous reward function. $\Bar{\Pr}$ and $\Bar{r}$ are respectively defined by 
\begin{align}
    \Bar{\Pr}(\Bar{s}', o|\Bar{s}, \gamma) &= \Pr(\Bar{s}', o|\Bar{s}, a)\quad\forall \Bar{s}, \Bar{s}'\in\Bar{\cS}, o\in\cO, \gamma\in\Bar{\cA} \\
    \Bar{r}(\Bar{s}, \gamma) &=  r(s, a)\quad\forall \Bar{s}\in\Bar{\cS}, \gamma\in\Bar{\cA}
\end{align}
where $\Bar{s} = (s, m), a=(a^i)_{i\in\cI}$ and $a^i = \gamma^i(m^i)$.

In this paper, we focus on the coordinator's POMDP and its variations. Without lost of generality, we remove the overline of $\Bar{\cS}$ and assume that the private information is a fixed function of the state. For $s\in\cS$, we use $m_s=(m_s^i)_{i\in\mathcal{I}}\in\cM$ to denote its corresponding private information. We also drop the overline of $\Bar{\Pr}$ and $\Bar{r}$ to simplify expressions. Note that we still use $\bcA$ to represent the space of prescriptions to distinguish it from the space of actions of individual agents. 

\begin{remark}
    The coordinator's POMDP can have a prohibitively large action space even for seemingly small problems like the DecTiger problem with $N\leq 3$ doors and 1-step delayed information sharing (described in Appendix \ref{dectiger}).
    For example, with $N=2$, we have $|\Bar{\mathcal{A}}| = \prod_{i=1}^2|\cA^i|^{|\cM^i|} = 3^{14} \approx 4.78 \times 10^6$; with $N=3$, we have $|\Bar{\mathcal{A}}| = \prod_{i=1}^2|\cA^i|^{|\cM^i|} = 4 ^ {26} \approx 4.50 \times 10^{15}$. For POMDPs with such extraordinarily large action space, most computers wouldn't even have enough memory to initialize off-the-shelf POMDP solvers, let alone running them.
\end{remark}

\subsection{Point-based Algorithms for POMDPs}\label{sec:prelim-pomdps}

In this section, we provide an overview of point-based algorithms for POMDPs and their common ingredient: the point-based backup operation. We then discuss the heuristic search value iteration (HSVI) algorithm \cite{hsvi}, which will be the basis of our proposed algorithm. To describe the algorithms in this section, we consider a standard single-agent infinite-horizon discounted reward POMDP defined through a tuple $\mathcal{E}=(\cS, \cA, \cO, b_0, \Pr, r, \beta)$.

\subsubsection{Point-based Algorithms and the Backup Operation}\label{sec:pointbasedbackup}
For a POMDP, the value function is defined as a function of the belief state and the Bellman operator is defined as follows: For any function $V:\Delta(\cS)\mapsto \mathbb{R}$,
\begin{equation}
    TV(b) := \max_{a\in\cA} \left[ r(b, a) + \beta\sum_{o\in\cO} \Pr(o|b, a) V(\tau(b, a, o)) \right]
\end{equation}
where $b \in \Delta(\cS) $ and $\tau$ is the belief update function. Since the value function $V$ is known to be piecewise-linear and convex, it can be written as $V(b) = \max_{\alpha\in \cV} \alpha^T b$  where $\cV$ is a finite collection of $|\cS|-$dimensional vectors.
\blue{
Then we have
\begin{align}
    &\quad~TV(b) \\
    &= \max_{a\in \cA} r(b, a) + \sum_{o\in\cO} \Pr(o|b, a) \max_{\alpha\in\cV} [\tau(b, a, o)]^T \alpha\\
    &= \max_{a\in \cA} r(b, a) + \sum_{o\in\cO} \max_{\alpha\in\cV}[\Pr(o|b, a) \tau(b, a, o)]^T \alpha\\
    &= \max_{a\in \cA} r(b, a) + \sum_{o\in\cO} \max_{\alpha\in\cV} \sum_{s'\in\cS} \Pr(s', o|b, a)\alpha(s')\\
    &= \max_{a\in \cA} \max_{\mu\in\cV^\cO } r(b, a) + \sum_{o\in\cO} \sum_{s'\in\cS} \Pr(s', o|b, a)\mu_o(s')\\
    &= \max_{a\in \cA} \max_{\mu\in \cV^{\cO}} \left(\alpha^{a, \mu}\right)^T b
\end{align}
} 
where $\cV^{\cO}$ is the set of all mappings from $\cO$ to $\cV$ and
\begin{equation}
    \alpha^{a, \mu}(s):= r(s, a) + \beta \sum_{o\in\cO}\sum_{s'\in\cS}\Pr(s', o|s, a)\mu_o(s').
\end{equation}

Therefore, the value function can be achieved by updating the collection of $\alpha$-vectors \cite{sondik1978optimal, cassandra1997incremental,shani2013survey} through 
\begin{align}
    \cV^{(k+1)} &:= \{\alpha^{a, \mu}: a\in\cA, \mu\in(\cV^{(k)})^\cO \}.\label{eq:exactvi}
\end{align}

However, computing \eqref{eq:exactvi} is  inefficient: We can see that $|\cV^{(k+1)}| = |\cA|\cdot|\cV^{(k)}|^{|\cO|}$, i.e. the growth of number of $\alpha$-vectors is doubly exponential. Even with procedures to prune out dominated $\alpha$-vectors \cite{cassandra1997incremental}, the exact value iteration is still ill-equipped to handle large state and action spaces \cite{shani2013survey}.

While computing all the $\alpha$-vectors in a Bellman update is expensive, computing an $\alpha$-vector that supports the updated function at a particular belief point $b\in\Delta(\cS)$ (i.e. $TV(b) = \alpha^T b$ and $TV(\Tilde{b}) \geq \alpha^T \Tilde{b}$ for all $\Tilde{b}\in\Delta(\cS)$) is not: one can compute this vector $\alpha^b$ through
\begin{align}
    \alpha^{b, a, o} &:= \argmax_{\alpha\in\cV} \sum_{s\in\cS}b(s)\sum_{s'\in\cS} \Pr(s', o|s, a)\alpha(s')\label{eq:alphabao}\\
    &\qquad\qquad\qquad\qquad\forall a\in \cA, o\in \cO\\
    \alpha^{b, a}(s) &:= r(s, a) + \beta \sum_{o\in\cO} \sum_{s'\in\cS} \Pr(s', o|s, a)\alpha^{b, a, o}(s')\label{eq:alphaba}\\
    &\qquad\qquad\qquad\qquad\forall s\in\cS, a\in\cA\\
    \alpha^b &:= \argmax_{\alpha^{b, a}: a\in\cA } \left(\alpha^{b, a}\right)^T b\label{eq:alphab}
\end{align}
\noeqref{eq:alphabao}\noeqref{eq:alphaba}\noeqref{eq:alphab}

The above procedure is referred to as the (point-based) \emph{backup} procedure at belief point $b$. Point-based algorithms for POMDPs are based on the idea that by using backup procedures at a carefully selected subset of belief points, one can obtain a set of $\alpha$-vectors that provide a good approximation of the optimal value function.

\begin{algorithm}[!ht]
    \SetKwFunction{FB}{Backup}
    \SetKwFunction{FPB}{GenericPointBased}
    \SetKwFunction{FOS}{DirectControlPolicy}
    \SetKwProg{Fn}{Function}{:}{end}

    \Fn{\FPB}{
    \KwIn{A single-agent POMDP model $(\cS, \cA, \cO, b_0, \Pr, r, \beta )$}
    \KwOut{A belief based policy $\pi$ and a lower bound of $V^{\pi}(b_0)$}

    \BlankLine
    Initialize $\cV$\;
    \While{Stopping criterion not satisfied}{
        Choose belief point set $\mathcal{B}$\;
        $\alpha^b = $ \FB{$\cV, b$} for all $b\in\mathcal{B}$\;
        Add $\{\alpha^b\}_{b\in\mathcal{B}}$ to $\cV$\;
        Prune dominated vectors in $\cV$ once in a while\;
    }
    $\pi := $ \FOS{$\cV$}\;
    $\underline{v} := \max_{\alpha\in\cV}\alpha^T b_0$\;
    \KwRet{$\pi, \underline{v}$}
    }

    \caption{A generic point-based algorithm for single-agent POMDP}\label{alg:genericpointbased}
\end{algorithm}
\blue{
Moreover, if the initial set $\cV$ represents a uniformly improvable (as defined in \cite{zhang2001speeding}) lower bound of the optimal value function, then $\cV$ will remain a uniformly improvable lower bound after inserting new $\alpha$-vectors from point-based updates. It can also be shown that \cite{hsvi2} in this case, the value of the direct control policy\footnote{The direct control policy \cite{hauskrecht2000value,hsvi2} can be described as follows: At belief $b$, find an $\alpha$-vector $\alpha^*\in\cV$ that maximizes $\alpha^T b$, and take the action $a^*$ associated with $\alpha^*$, i.e. the maximizing action in \eqref{eq:alphab} when $\alpha^*$ was initially computed.} obtained from $\cV$ is lower bounded by $V$, the piece-wise linear convex function with $\alpha$-vectors $\cV$. Therefore we have a theoretical guarantee of the resulting policy. A generic framework for many point-based algorithms (e.g. \cite{perseus,hsvi,sarsop,gapmin}) is shown in Algorithm \ref{alg:genericpointbased}. These algorithms all share the same backup procedure. They differ mostly in how they determine the set of belief points to perform backups on.
} 

\subsubsection{Heuristic Search Value Iteration}\label{sec:hsvi}
\blue{
The HSVI algorithm in \cite{hsvi} is summarized in Algorithm \ref{alg:hsvi}. 
In addition to a set of $\alpha$-vectors that represent a lower bound for the optimal value function $V^*$,
the algorithm maintains an upper bound of $V^*$ as well. At each time, the algorithm performs a depth-first search to select a few belief points. The algorithm updates both the upper bound and lower bound at those beliefs. In those updates, the algorithm computes exact Bellman updates of the upper and lower bounds at those beliefs, and incorporate the new values back into the bounds. As a result, the lower bound increases and upper bound decreases over time. HSVI is an anytime algorithm that always tries to shrink the gap between the upper and the lower bound. Once the algorithm terminates once a certain stopping criteria is met, it returns the direct control policy associated with the $\alpha$-vectors used for the lower bound function.

The HSVI algorithm distinguishes itself from previous algorithms through its use of an upper-bound based heuristic search procedure, as it is defined in the function \texttt{ChooseNext}. To determine the next belief to explore, the algorithm first optimistically chooses an action that maximizes the action-value function (or Q-function) associated with the upper bound. Then, the algorithm picks an observation that maximizes the \emph{excess gap}. This procedure allows the algorithm to focus on the beliefs where the current gap between the upper bound and lower bound is unsatisfactory.} 

The lower bound function $L$ in HSVI is represented as a set of $\alpha$-vectors. Its update function $L$.\texttt{Update}($b$) adds a new $\alpha$-vector to the set by performing the backup operation at $b$. The algorithm also periodically prunes certain dominated $\alpha$-vectors in the set.

Unlike the lower bound, the upper bound function in HSVI is represented through the lower convex hull of a set of isolated points $(b, \Bar{v}_b) \in \Delta(\cS)\times \mathbb{R}$, i.e., let $\mathbf{B}$ be a matrix whose $n$ column vectors represent $n$ beliefs and $\Bar{v}\in\mathbb{R}^n$ be a vector representing the upper bound values of those beliefs, then for any $b \in \Delta(\cS), $
\begin{equation}\label{eq:ublp}
    U(b) = \min \{ \Bar{v}^T \eta: \eta\in \mathbb{R}_+^{n}, \mathbf{B}\eta = b. \}
\end{equation}
This representation is based on the fact that $V^*$ is convex. 
Let $T$ be the Bellman operator as defined in Section \ref{sec:pointbasedbackup}. The $U$.\texttt{Update}($b$) procedure in Algorithm \ref{alg:hsvi} computes 
\begin{equation}\label{eq:bellmanub}
    \Bar{v}_b := TU(b) = \max_{a\in\cA} r(b, a) + \beta\sum_{o\in\cO} \Pr(o|b, a) U(\tau(b, a, o))
\end{equation}
and then adds $(b, \Bar{v}_b)$ to the point set used for convex hull. The algorithm also removes redundant points in the set periodically. The procedure returns $a^*\in \cA$, an optimizer of \eqref{eq:bellmanub}, for potential future use in the search heuristics.

\begin{algorithm}[!ht]
    \SetKwFunction{FHSVI}{HSVI}
    \SetKwFunction{Fexplore}{Explore}
    \SetKwFunction{FOS}{DirectControlPolicy}
    \SetKwFunction{FU}{Update}
    \SetKwFunction{FB}{Backup}
    \SetKwFunction{Ac}{Access}
    \SetKwFunction{CN}{ChooseNext}
    \SetKwProg{Fn}{Function}{:}{end}
    \SetKwProg{Cl}{Class}{:}{end}
    \SetKw{And}{and}
    \Fn{\FHSVI}{   
        \KwIn{A single-agent POMDP model $(\cS, \cA, \cO, b_0, \Pr, r, \beta )$}
        \KwOut{A belief based policy $\pi$; an upper bound for $V^*(b_0)$ and lower bound for $V^\pi(b_0)$.}
        \BlankLine
        \tcp{$\zeta\in(0, 1)$ is a hyperparameter.}
        Initialize $U$ and $L$\;
        \While{Stopping criterion not satisfied}{
            $\epsilon = \zeta [U(b_0) - L(b_0)]$\;
            \Fexplore{$U, L, b_0, \epsilon$}\;     
        }
        $\pi = ~$\FOS{$L$}\;
        \KwRet{$\pi, U(b_0), L(b_0)$}
    }

    \BlankLine
    \Fn{\Fexplore{$U, L, b, \epsilon$}}{
        \tcp{Modifies $U, L$}
        $a^* = U$.\FU{$b$}\;
        $L$.\FU{$b$}\;
        \lIf{$U(b) - L(b) \leq \epsilon$}{\KwRet}
        $(b', \epsilon') = $ \CN{$U, L, b, a^*, \epsilon$}\;
        \Fexplore{$U, L, b', \epsilon'$}\;
        $U$.\FU{$b$}\;
        $L$.\FU{$b$}\;
    }

    \BlankLine
    \Fn{\CN{$U, L, b, a^*, \epsilon$}}{
        $o^* = \argmax_{o\in\cO} \Pr(o|b, a^*)[U(\tau(b, a^*, o)) - L(\tau(b, a^*, o)) - \frac{\epsilon}{\beta})]$\;
        \KwRet{$\tau(b, a^*, o^*), \frac{\epsilon}{\beta}$}\;
    }
    
    \caption{Heuristic Search Value Iteration}\label{alg:hsvi}
\end{algorithm}



\section{Coordinator's HSVI Algorithm}
\label{sec:coord-hsvi}

As noted earlier, compared to a regular single-agent POMDP, a coordinator's POMDP will have an exponentially large number of actions. This makes both the lower bound and upper bound update of HSVI infeasible. More specifically, in the backup operations \eqref{eq:alphabao}-\eqref{eq:alphab}, we need to first solve $|\Bar{\cA}| \times |\cO|$ discrete optimization problems in \eqref{eq:alphabao}  and then solve an optimization problem over $\Bar{\cA}$ in \eqref{eq:alphab}. Also, to perform a Bellman update for the upper bound at belief point $b$, the HSVI algorithm needs to compute $U(\tau(b, \gamma, o))$ for all pairs of $(\gamma, o)\in\Bar{\cA}\times \cO$. This is infeasible also due to the large number of actions in a coordinator's POMDP.

In this section, we introduce the \emph{Coordinator's Heuristic Search Value Iteration} (CHSVI) algorithm, which combines the CI approach with the HSVI algorithm. Instead of applying HSVI directly on the coordinator's POMDP, we apply HSVI on a multi-step extended form of coordinator's POMDP. Through  the use of the extended form and the structure of prescription space, we are able to simplify both the upper bound and lower bound update operations, creating a more practical algorithm.

We first describe the  extended form of coordinator's POMDP. For the ease of illustration, consider $\mathcal{I} = \{1, 2\}$ though the idea can naturally extend to more than two players. We derive an extended POMDP $\hat{\mathcal{E}}$ from the coordinator's POMDP $\Bar{\mathcal{E}}=\{\cS, \Bar{\cA}, \cO, b_0, \mathbb{P}, r, \beta\}$ by extending each time $t$ into three stages: $(t, 0), (t, 1), (t, 2)$. The state space is $\cS^0:=\cS$ at $(t, 0)$, $\cS^1 :=\cS\times \cA^1$ at $(t, 1)$, and $\cS^2:=\cS\times \cA$ at $(t, 2)$. The common observation space is $\{\emptyset\}$ at both $(t, 0)$ and $(t, 1)$, and $\cO$ at $(t, 2)$. Only stage $(t,2)$ features an instantaneous reward, which is $r(s, a)$ for $(s, a)\in\cS^{2}$. The system evolves as follows:

\begin{enumerate}
    \item At $(t, 0)$, the coordinator chooses prescription $\gamma^1 \in \Bar{\cA}^1$ for agent 1 only. The state deterministically transits from $s\in\cS$ to $(s, \gamma^1(m_s^1))\in\cS\times\cA^1$. 
    
    \item At $(t, 1)$, the coordinator chooses prescription $\gamma^2 \in \Bar{\cA}^2$ for agent 2 only. The state deterministically transits from $(s, a^1)\in\cS\times\cA^1$ to $(s, a^1, \gamma^2(m_s^2))\in\cS\times\cA^1\times\cA^2$. 
    
    \item At $(t, 2)$, the coordinator has no action to take. The state-information joint transition kernel is simply the same $\Pr(s', o|s, a)$ as defined in the initial model $\mathcal{E}$.
\end{enumerate}

The total reward for the new POMDP is given by $\sum_{t=0}^\infty \beta^t r(s_{(t, 2)}, a_{(t, 2)})$. With some abuse of notation, for $\gamma^i\in\Bar{\cA}^i$, we define $\gamma^i(a^i|m^i) := \bm{1}_{\{\gamma^i(m^i) = a^i\} }$. The belief update functions for the three stages are given by
\begin{align}
    &[\tau^0(b, \gamma^1)](s, a^1) = b(s)\gamma^1(a^1|m_s^1)\\
    &[\tau^1(b, \gamma^2)](s, a^1, a^2) = b(s, a^1)\gamma^2(a^2|m_s^2)\\
    &[\tau^2(b, o)](s') = \dfrac{\sum_{(s, a)\in\cS\times\cA}\Pr(s', o|s, a)b(s, a)}{\sum_{\Tilde{s}\in\cS} \sum_{(s, a)\in\cS\times\cA} \Pr(\Tilde{s}, o|s, a)b(s, a) }
\end{align}

The extended coordinator's POMDP $\hat{\mathcal{E}}$ differs from the coordinator's POMDP $\Bar{\mathcal{E}}$ only in the ordering of events within a time instant $t$ but not in the total reward at time  $t$ and the dynamics from $t$ to $t+1$. 
Therefore, $\hat{\mathcal{E}}$ is equivalent to  $\Bar{\mathcal{E}}$: Any strategy in $\hat{\mathcal{E}}$ has a counterpart in $\Bar{\mathcal{E}}$ with the same total reward and vice versa. 

In CHSVI algorithm (Algorithm \ref{alg:chsvi}), we apply the framework of HSVI to the extended coordinator's POMDP while utilizing the special structure of the prescription space to implement the update procedures more efficiently. 
In the following sections, we describe the upper bound $U$ and lower bound $L$ used in the CHSVI algorithm in detail.

\blue{
\begin{remark}
    In this section we have taken an agnostic approach to transform any coordinator's POMDP into an extended coordinator's POMDP, where the intermediate state spaces are $\cS\times \cA^1$ and $\cS\times\cA^1\times\cA^2$, and only one agent (including nature) takes action at one stage. In fact, the idea of the CHSVI algorithm can be applied to any transformation of a coordinator's POMDP such that only one agent takes action at one stage.
\end{remark}
} 

\begin{algorithm}[!ht]
    \SetKwFunction{FCHSVI}{CHSVI}
    \SetKwFunction{Fexplore}{Explore}
    \SetKwFunction{FOS}{DirectControlPolicy}
    \SetKwFunction{FU}{Update}
    \SetKwFunction{FB}{Backup}
    \SetKwFunction{Ac}{Access}
    \SetKwFunction{CN}{ChooseNext}
    \SetKwProg{Fn}{Function}{:}{end}
    \SetKwProg{Cl}{Class}{:}{end}
    \SetKw{And}{and}

    \Fn{\FCHSVI}{
        \KwIn{A coordinator's POMDP model $(\cS, \Bar{\cA}, \cO, b_0, \Pr, r, \beta )$}
        \KwOut{A belief based coordination policy $\pi$; an upper bound for $V^*(b_0)$ and lower bound for $V^\pi(b_0)$.}

        \BlankLine
        \tcp{$\zeta\in(0, 1)$ is a hyperparameter.}
        Initialize $U$ and $L$\;
        \While{Stopping criterion not satisfied}{
            $\epsilon = \zeta [U(b_0) - L(b_0)]$\;
            \Fexplore{$U, L, b_0, \epsilon$}\;     
        }
        $\pi = ~$\FOS{$L$}\;
        \KwRet{$\pi, U(b_0), L(b_0)$}
    }

    \BlankLine
    \Fn{\Fexplore{$U, L, b, \epsilon$}}{
        $\gamma^* = U$.\FU{$b$}\;
        $L$.\FU{$b$}\;
        \lIf{$U(b) - L(b) \leq \epsilon$}{\KwRet}
        $(b', \epsilon') = $ \CN{$U, L, b, \gamma^*, \epsilon$}\;
        \Fexplore{$U, L, b', \epsilon'$}\;
        $U$.\FU{$b$}\;
        $L$.\FU{$b$}\;
    }

    \BlankLine
    \Fn{\CN{$U, L, b, \gamma^*, \epsilon$}}{
        Set $\ell\in\{0, 1, 2\}$ to be such that $b\in\Delta(\cS^\ell)$\;
        \uIf{$\ell < 2$}{
        \KwRet{$\tau^{\ell}(b, \gamma^*), \epsilon$}\;
        }
        \Else{
        $o^* = \argmax_{o\in\cO} \Pr(o|b)[U(\tau^2(b, o)) - L(\tau^2(b, o)) - \epsilon\beta^{-1}]$\;
        \KwRet{$\tau^2(b, o^*), \epsilon\beta^{-1}$}\;
        }
    }
    \caption{Coordinator's HSVI Algorithm}\label{alg:chsvi}
\end{algorithm}

\subsection{Lower Bound Update}
In CHSVI, we maintain three lower bound functions (sets of $\alpha$-vectors) corresponding to the three stages of $t$. Let $L^{\ell}$ denote the lower bound function and $\mathcal{V}^{\ell}$ denote the corresponding set of $\alpha$-vectors for stage $\ell$ (of some time $t$).  Let $T^\ell$ be the Bellman operator for stage $\ell$. We now describe the update steps in detail.

\medskip
\noindent\textbf{Stage} $\ell = 0, 1$: In this stage, the coordinator picks a prescription for agent $i=\ell+1$. For $b\in\Delta(\cS^\ell)$, the Bellman update of the lower bound at $b$ is given by
    \begin{align}
        &\quad~[T^{\ell}L^{\ell+1}](b) = \max_{\gamma^i\in\Bar{\cA}^i} L^{\ell+1}(\tau^\ell(b, \gamma^i))\\
        &= \max_{\gamma^i\in\Bar{\cA}^i}\max_{\alpha\in \cV^i} \sum_{(s^\ell, a^i)\in\cS^\ell\times\cA^i}b(s^\ell) \gamma^i(a^i|m_{s^\ell}^i) \alpha(s^\ell, a^i)\quad\label{eq:backupt0}
    \end{align}

    \noeqref{eq:chsvi:lb:1}\noeqref{eq:chsvi:lb:2}\noeqref{eq:chsvi:lb:3}\noeqref{eq:chsvi:lb:4}
    The optimization problem \eqref{eq:backupt0} can be solved via the following steps:
    \begin{align}
        \gamma^{i, \alpha}(m^i) &:= \argmax_{a^i\in\cA^i} \sum_{\substack{s^\ell\in\cS^\ell: m_{s^\ell}^i = m^i}} b(s^\ell)\alpha(s^\ell, a^i)\label{eq:chsvi:lb:1}\\
        &\qquad\qquad\quad\forall m^i\in\cM^i, \alpha\in\cV^i \\
        J(\alpha)&:=\sum_{s^\ell\in\cS^\ell}b(s^\ell)\alpha(s^\ell, \gamma^{i, \alpha}(m_{s^\ell}^i) )\quad\forall \alpha\in\cV^i\label{eq:chsvi:lb:2}\\
        \alpha^* &:= \argmax_{\alpha\in\cV^i} J(\alpha)\label{eq:chsvi:lb:3}\\
        \gamma^{i, *} &:= \gamma^{i, \alpha^*},\label{eq:chsvi:lb:4}
    \end{align}
    where we have used the fact that for each fixed $\alpha\in\cV^i$, the optimization problem over $\gamma^i\in\Bar{\cA}^i$ can be separated into $|\cM^i|$-optimization problems over $\cA^i$. Then, we add the following new alpha vector $\alpha^b\in\mathbb{R}^{\cS^\ell}$ to $\cV^\ell$:
    \begin{equation}\label{eq:chsvi:lb:alphab}
        \alpha^b(s^\ell) = \sum_{a^i\in\cA^i}\gamma^{i, *}(a^i|m_{s^\ell}^i) \alpha^*(s^\ell, a^i)\quad\forall s^\ell\in\cS^\ell.
    \end{equation}

    \begin{remark}
        If we apply the point-based backup procedure \eqref{eq:alphabao} -- \eqref{eq:alphab} directly at this stage, then the operation count would be $\Theta(|\cS| |\cV| |\Bar{\cA}^i| )$, which grows exponentially in $|\cM^i|$ since $|\Bar{\cA}^i| = |\cA^i|^{|\cM^i|}$. In contrast, the operation count of the procedure listed in \eqref{eq:chsvi:lb:1} -- \eqref{eq:chsvi:lb:alphab}  is $\Theta(|\cS| |\cV^i| |\cA^i| |\cM^i|)$, which is polynomial in all parameters involved. 
    \end{remark}

\blue{
    \begin{remark}
        We have used the separability of the optimization problem over the prescription space $\Bar{\cA}^i$ for a fixed $\alpha$-vector to simplify the optimization problem \eqref{eq:backupt0}. This can only be achieved if there's only one agent at one stage, which is part of the reason why we use the extended form of coordinator's POMDP. 
    \end{remark}
} 

\noindent\textbf{Stage} $2$: For $b\in\Delta(\cS^2)$ the Bellman update of the lower bound at $b$ is given by
    \begin{align*}
        &[T^2L^0](b) = r(b) + \beta\sum_{o\in\cO} \Pr(o|b) \max_{\alpha\in\cV^0}  [\tau^2(b, o)]^T \alpha\\
        &= r(b) + \beta\sum_{o\in\cO} \max_{\alpha\in\cV^0} \sum_{s^2\in\cS^2}b(s^2)\sum_{s'\in\cS} \Pr(s', o|s^2) \alpha(s')
    \end{align*}
    where $r(b):=\sum_{(s, a)\in\cS\times\cA}b(s, a)r(s, a)$.

    Since we  have no action in this stage, we can compute the new alpha vector $\alpha^b\in\mathbb{R}^{\cS^2}$ in the same way as in \eqref{eq:alphabao}-\eqref{eq:alphab} (except that there's no action), i.e. we compute
    \begin{align}
        \alpha^{b, o} &:= \argmax_{\alpha\in \cV^0} \sum_{s^2\in\cS^2}b(s^2)\sum_{s'\in\cS} \Pr(s', o|s^2) \alpha(s')\\
        &\qquad\qquad\qquad\qquad\qquad\qquad\quad\forall o\in\cO\label{eq:chsvi:lb:alphabo}\\
        \alpha^b(s^2) &:= r(s^2) + \beta\sum_{o\in\cO}\sum_{s'\in\cS} \Pr(s', o|s^2)\alpha^{b, o}(s')\\
        &\qquad\qquad\qquad\qquad\qquad\qquad\quad\forall s^2\in\cS^2 \label{eq:chsvi:lb:alphab2}
    \end{align}

The lower bound update algorithm is given in Algo. \ref{alg:chsvi:lbupdate}.
\begin{algorithm}
    \SetKwFunction{FU}{Update}
    \SetKwProg{Fn}{Function}{:}{end}

    \Fn{$L$.\FU{$b$}}{
        \tcp{$L$ stores three sets of $\alpha$-vectors: $\cV^{\ell}, \ell=0,1,2$ }
        Set $\ell\in\{0, 1, 2\}$ to be such that $b\in\Delta(\cS^\ell)$\;
        \uIf{$\ell<2$}{
            Compute $\alpha^b$ with \eqref{eq:chsvi:lb:1} --  \eqref{eq:chsvi:lb:alphab}\;
        }
        \Else{
            Compute $\alpha^b$ with \eqref{eq:chsvi:lb:alphabo}\eqref{eq:chsvi:lb:alphab2}\;
        }
        Add $\alpha^b$ to $\cV^\ell$\;
        Prune dominated vectors in $\cV^\ell$ once in a while\;
    }
    \caption{Lower Bound Update}\label{alg:chsvi:lbupdate}
\end{algorithm}

\subsection{A New Upper Bound Representation}\label{sec:ubrepr}
In this section, we propose a new upper bound representation for POMDPs, called $\alpha$-constraints based upper bound. 
\blue{Our motivation for finding a new upper bound representation comes from a weakness of the convex hull-based upper bound in extended coordinator's POMDPs. We believe that our new upper bound can alleviate the weakness. In any case, the new upper bound is never worse (i.e. larger) than the original upper bound.

In HSVI, the upper bound function plays a very important role in the search heuristics. It is represented with a collection of isolated points $(b, v^b)\in\Delta(\cS)\times\mathbb{R}$. The upper bounds at other belief points are obtained via convex hull based interpolation. 
While convex hull based upper bounds and their approximations have achieved empirical success \cite{hsvi2,sarsop} on many single-agent POMDP problems, such bounds can be insufficient for the extended coordinator's POMDP due to the following reason: For two probability distributions $b$ and $\Tilde{b}$ defined on the same space, we say that $b$ is \emph{inexpressible} with $\Tilde{b}$ if in any convex combination that represents $b$, $\Tilde{b}$ is not involved, i.e. its coefficient is 0. The inclusion of actions into the state space creates a exponentially large number of mutually inexpressible beliefs. As a result, the upper bound at those beliefs do not improve (i.e. become smaller) from their initial values unless it is explored by the algorithm. Due to the optimism-based exploration heuristics of these algorithms \cite{hsvi2,sarsop}, it is likely that the algorithm would end up exhaustively sampling those beliefs, resulting in a behavior similar to brute-force search.

To illustrate this point more clearly, consider a belief $b_0\in\Delta(\cS)$ with full support. 
Then, consider two prescriptions $\gamma^1, \Tilde{\gamma}^i\in\Bar{\cA}^1$. Suppose that $\gamma^1(m^1)=a^1\neq \Tilde{\gamma}^1(m^1)=\Tilde{a}^1$ for some $m^1\in \cM^1$. Let $b_1 = \tau^0(b_0, \gamma^1)$ and $\Tilde{b}_1 = \tau^0(b_1, \Tilde{\gamma}^1)$. Let $s\in\cS$ be some state such that $m_s^1 = m^1$. We have $b_1(s, a^1) > 0, \Tilde{b}_1(s, a^1) = 0$. This means that $\Tilde{b}_1$ is inexpressible with $b_1$ and vice versa. As a result, updating the upper bound at $b_1$ yields no effect on the upper bound at $\Tilde{b}_1$, even if $b_1$ is very close to $\Tilde{b}_1$ (which is the case when $\gamma^1$ differs from $\Tilde{\gamma}^1$ only at one $m^1\in\cM^1$).

To resolve this problem, we would like an upper bound representation that is not only tight, but also facilitates more \emph{cross learning}, i.e. knowing the upper bound of a belief $b$ helps us to reduce the upper bound another belief $\Tilde{b}$, even if $\Tilde{b}$ is inexpressible with $b$. Furthermore, if $b$ and $\Tilde{b}$ is close, then we would like their upper bound to be close as well. 

To design a new upper bound representation, we start from the dual form of the convex hull representation \eqref{eq:ublp} given by} 
\begin{equation}\label{eq:ublpdual}
\begin{split}
    U(b) = \max \{ b^T y: y\in \mathbb{R}^{\cS}, \mathbf{B}^T y  \leq \Bar{v} \}.
\end{split}
\end{equation}

A direct interpretation of \eqref{eq:ublpdual} is as follows: The optimal value function has an $\alpha$-vector representation $V^*(b) = \max_{\alpha\in\cV^*} \alpha^T b$, where $\cV^*$ represents the limit of $\cV^{(k)}$ defined in \eqref{eq:exactvi}. The maximization problem \eqref{eq:ublpdual} provides an upper bound for $V^*(b)$ since all $\alpha$-vectors in $\cV^*$ satisfy the constraints $\mathbf{B}^T \alpha \leq \Bar{v}$ (due to the fact that $\Bar{v}$ is an upper bound for $V^*$ at beliefs in $\mathbf{B}$). In other words, \eqref{eq:ublpdual} can be seen as an $\alpha$-\emph{constraint} based upper bound, where the upper bound function is constructed from a group of linear constraints the set $\cV^*$ should satisfy.

Building upon this idea, we make use of other types of linear constraints other than value function upper bounds, e.g., $\alpha(s) \geq v_{\min}$ where $v_{\min} := \min_{s\in\cS,a\in\cA}r(s, a) / (1-\beta)$. Adding this to \eqref{eq:ublpdual}, we have
\begin{equation}\label{eq:alpconstrub}
\begin{split}
    U(b) = \max \{ b^T y: y\in \mathbb{R}^{\cS}, 
        \mathbf{B}^T y  \leq \Bar{v}, y \geq v_{\min} \bm{1} \}.
\end{split}
\end{equation}
to be an upper bound no worse than the original one used in HSVI. 
\blue{
In contrast to \eqref{eq:ublpdual}, the upper bound \eqref{eq:alpconstrub} can facilitate more cross learning: By bounding the $L_\infty$ diameter of the constraint set, it can be shown that \eqref{eq:alpconstrub} is $L_1$-Lipschitz continuous with constant $v_{\max} - v_{\min}$ (where $v_{\max} := \max_{s\in\cS,a\in\cA}r(s, a) / (1-\beta)$), hence ensuring that if $b$ is close to $\Tilde{b}$, then their upper bounds are relatively close as well, no matter whether $\Tilde{b}$ is expressible with $b$ or not.} 
See Appendix \ref{app:morealphaconstr} for more types of $\alpha$-constraints.

In summary, any valid linear inequalities that are satisfied by all vectors in $\cV^*$ can be added to the set of $\alpha$-constraints. This provides us a lot of flexibility compared to the convex hull-based bounds.
We next describe the update of $\alpha$-constraint based upper bound in extended coordinator's POMDPs. The update method described in the next section is independent of the specific choice of $\alpha$-constraints.

\subsection{Upper Bound Update}
We maintain three upper bound functions corresponding to the three stages. Each upper bound function is represented through a set of $\alpha$-constraints as described in the previous section. Let $\mathbf{M}^{\ell}y\leq w^{\ell}$ represent the $\alpha$-constraints (on $y\in\cS^\ell$) associated with stage $\ell$. Let $T^\ell$ be the Bellman operator for stage $\ell$. We now describe the update steps in detail:

\smallskip
\noindent\textbf{Stage} $\ell = 0, 1$: In this stage, the coordinator picks a prescription for agent $i=\ell+1$. For $b\in\Delta(\cS^\ell)$, the Bellman updated upper bound at $b$ is given by
    \begin{align}
        \Bar{v}^{b}&:=[T^\ell U^{\ell+1}](b) =\max_{\gamma^i\in\Bar{\cA}^i} U^{\ell+1}(\tau^\ell(b, \gamma^i) ) \\
        &=\max_{\gamma^i\in\Bar{\cA}^i}\max_{\substack{y\in \mathbb{R}^{\cS^i} \\ \mathbf{M}^i y \leq w^i }} \sum_{(s^\ell, a^i)\in\cS^\ell\times \cA^i}b(s^\ell) \gamma^i(a^i|m_{s^\ell}^i) y(s^\ell, a^i)\\
        &=:\max_{\gamma^i\in\Bar{\cA}^i}\max_{\substack{y\in \mathbb{R}^{\cS^i} \\ \mathbf{M}^i y \leq w^i }} J^i(b, \gamma^i, y) \label{eq:updatet0}
    \end{align}

    Now, notice that if we treat each $\gamma^i$ as a 0-1 indicator vector in $[0, 1]^{\cA^i\times\cM^i}$, then $J^i(b, \gamma^i, y)$ is \emph{bilinear} in $\gamma^i$ and $y$: It is linear in $\gamma^i$ for each fixed $y$ and linear in $y$ for each fixed $\gamma^i$. Therefore, \eqref{eq:updatet0} is a \emph{bilinear programming} (BP) problem, which has been studied extensively in the optimization 
    literature \cite{gallo1977bilinear,konno1976cutting,cogill2006approximation} 
    as well as decentralized control literature. \cite{petrik2007average,petrik2009bilinear}. 
    Notably, in \cite{petrik2007average}, the author provided a method to convert bilinear programs to MILPs\footnote{The method only applies to BP where the constraints on two groups of variables are separate. The optimization problem of \eqref{eq:updatet0} does lie in this category.}.
    Gurobi Optimization\textsuperscript{\texttrademark} has also developed specialized solvers for bilinear programs \cite{gurobibilinear}. 

\begin{remark}
    Even though bilinear programming problems are NP-hard \cite{tsitsiklis1985complexity} in general, modeling the upper bound update through BP still offers several advantages over the original update method in HSVI (i.e. separately solving the inner linear program in \eqref{eq:updatet0} for each $\gamma^i\in\Bar{\cA}^i$) for the following reasons: 
    (i) It allow us to apply systematic methods for BP to avoid brute-force enumeration of prescriptions. 
    (ii) Not all BPs fall into the NP-hard category.
    (iii) It opens up the door for approximate upper bound methods (e.g. certain relaxation of the MILP reformulation \cite{petrik2007average} of BP).
\end{remark}

\smallskip
\noindent\textbf{Stage} 2: For $b\in\Delta(\cS^2)$, the Bellman update of the upper bound at belief $b$ is given by
    \begin{align}
        \Bar{v}^{b}&= [T^2U^0](b) \\
        &= r(b) + \beta\sum_{o\in\cO} \Pr(o|b)\max_{\substack{y\in\mathbb{R}^\cS \\ \mathbf{M}^0 y \leq w^0 } } [\tau^2(b, o)]^T y \label{eq:updatet2}
    \end{align}
    which can be computed by solving $|\cO|$ linear programs.

The upper bound update algorithm is given in Algo. \ref{alg:chsvi:ubupdate}.

\begin{algorithm}
    \SetKwFunction{FU}{Update}
    \SetKwProg{Fn}{Function}{:}{end}

    \Fn{$U$.\FU{$b$}}{
        \tcp{$U$ stores three sets of $\alpha$-constraints: $\mathcal{C}^{\ell}, \ell=0,1,2$ }
        Set $\ell\in\{0, 1, 2\}$ to be such that $b\in\Delta(\cS^\ell)$\;
        \uIf{$\ell<2$}{
            Compute $\Bar{v}^b, \gamma^{i, *}$, the optimal value and optimizer of the bilinear program \eqref{eq:updatet0}\;
            
        }
        \Else{
            Compute $\Bar{v}^b$ with \eqref{eq:updatet2}\;
            Set $\gamma^{i, *}$ to represent the null prescription\;
        }
        Add $b^T y \leq \Bar{v}^b$ to $\mathcal{C}^\ell$\;
        Prune redundant $\alpha$-constraints once in a while\;
        \KwRet{$\gamma^{i, *}$}\;
    }
    \caption{Upper Bound Update}\label{alg:chsvi:ubupdate}
\end{algorithm}

\section{Experimental Results}\label{sec:experimental}

We implemented the CHSVI algorithm in Python (\url{https://github.com/dwtang/chsvi}). All BP and LP involved in the algorithm are solved with Gurobi Optimization Studio\textsuperscript{\texttrademark}. The hyperparameter $\zeta$ is set to $0.85$. We terminate the algorithm when the gap between the upper and lower bounds is less than 0.01 or if the run time has exceeded 24 hours. The lower bounds are initialized through the fixed action bound in the same way as in \cite{hsvi2}. The $\alpha$-constraints used in upper bounds are initialized to be the marginal belief based constraints (see Appendix \ref{app:mbbalpha}) obtained from a relaxed POMDP problem where all private information are assumed to be common. We adapt the same pruning strategy as HSVI: For lower bounds, we only prune $\alpha$-vectors that are pointwise dominated by another vector. For upper bounds, we remove redundant $\alpha$-constraints through solving linear programs.

We run the algorithm on \texttt{DecTiger} and \texttt{MultiCast} instances defined in Appendix \ref{dectiger} and \ref{multicast} respectively. The experiments are conducted on a computer with Intel Xeon\textsuperscript{\textregistered} E3-1231 v3 CPU (4 cores, 3.4Ghz) and 16GB RAM. The results are shown in Table \ref{tab:dectiger} and Figure \ref{fig:dectiger}, with additional plots in Appendix \ref{app:experiment}.
In the all instances except \texttt{DecTiger}(3,1,0.99), the algorithm is able to close the gap between the upper and lower bound and find a near optimal strategy. To the best of our knowledge, except when analytical solutions are available, these are the first provably optimal solutions for infinite-horizon multi-agent control problems.

\begin{table}[!ht]
    \centering
    \begin{tabular}{c|cccc}
         & $L(b_0)$ & $U(b_0)$ & Time (s) \\\hline
        \texttt{DecTiger}~(2,1,0.9)& 32.7704 & 32.7792 & 56 \\
        \texttt{DecTiger}~(2,1,0.99)& 388.4035 & 388.4134 & 2081 \\
        \texttt{DecTiger}~(3,1,0.9)& 6.7139 & 6.7236 & 63781 \\
        \texttt{DecTiger}~(3,1,0.99)& 76.2553 & 222.2532 & 86483 \\
        \texttt{MultiCast}~(8,8,0.1,0.2,0.9)&-4.8550 & -4.8450 & 346\\
        \texttt{MultiCast}~(16,16,0.2,0.4,0.9)&-10.0653 & -10.0400 & 86423
    \end{tabular}
    \caption{Experimental Results.}
    \label{tab:expresult}
\end{table}

\begin{figure}[!ht]
    \centering
    \includegraphics[width=0.49\textwidth]{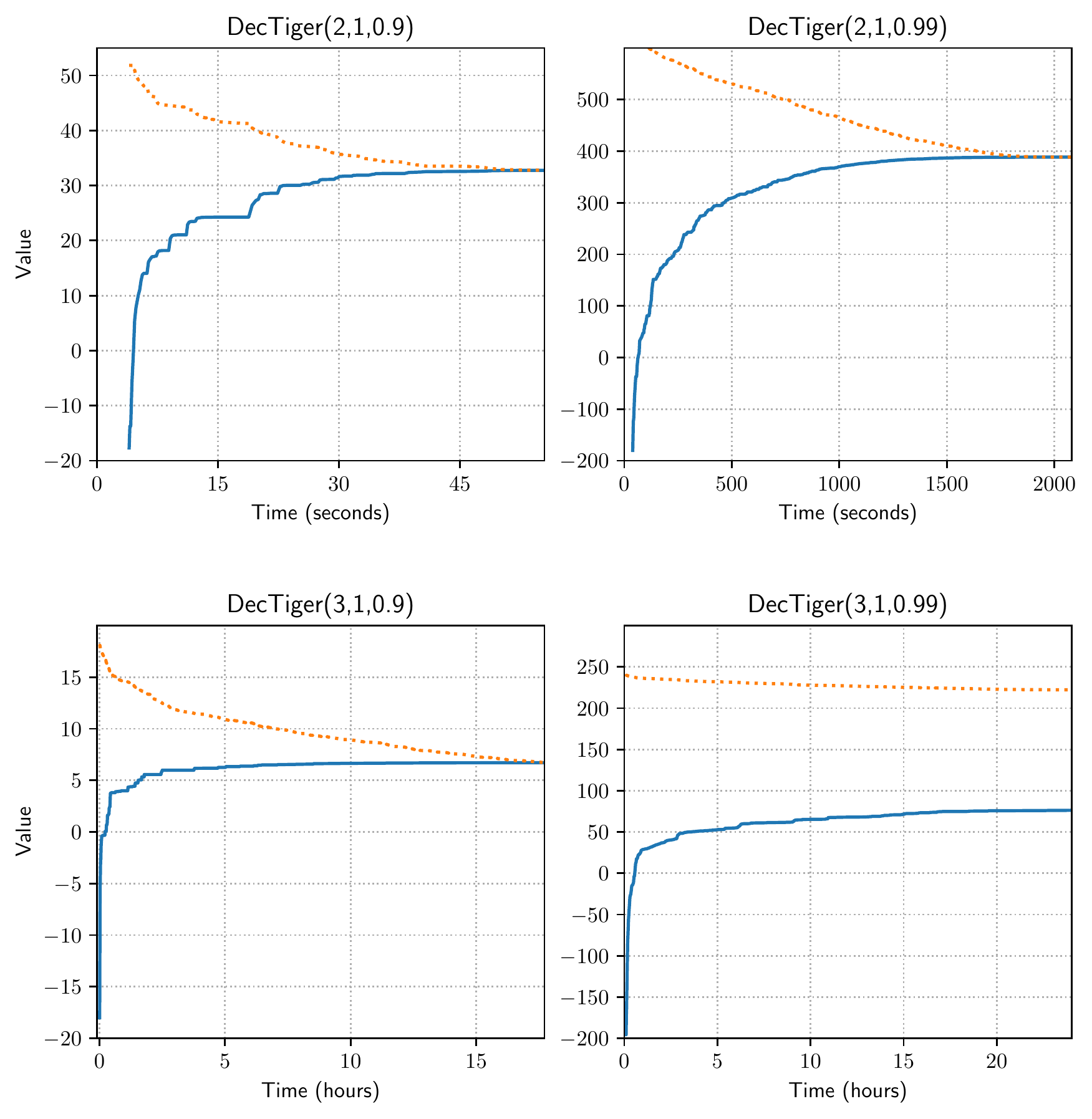}
    \caption{Experimental Results on \texttt{DecTiger}~($N, d, \beta$). The dotted line represents the upper bound $U(b_0)$ and the solid line represents the lower bound $L(b_0)$. }
    \label{fig:dectiger}
\end{figure}

\begin{figure}[!ht]
    \centering
    \includegraphics[width=0.49\textwidth]{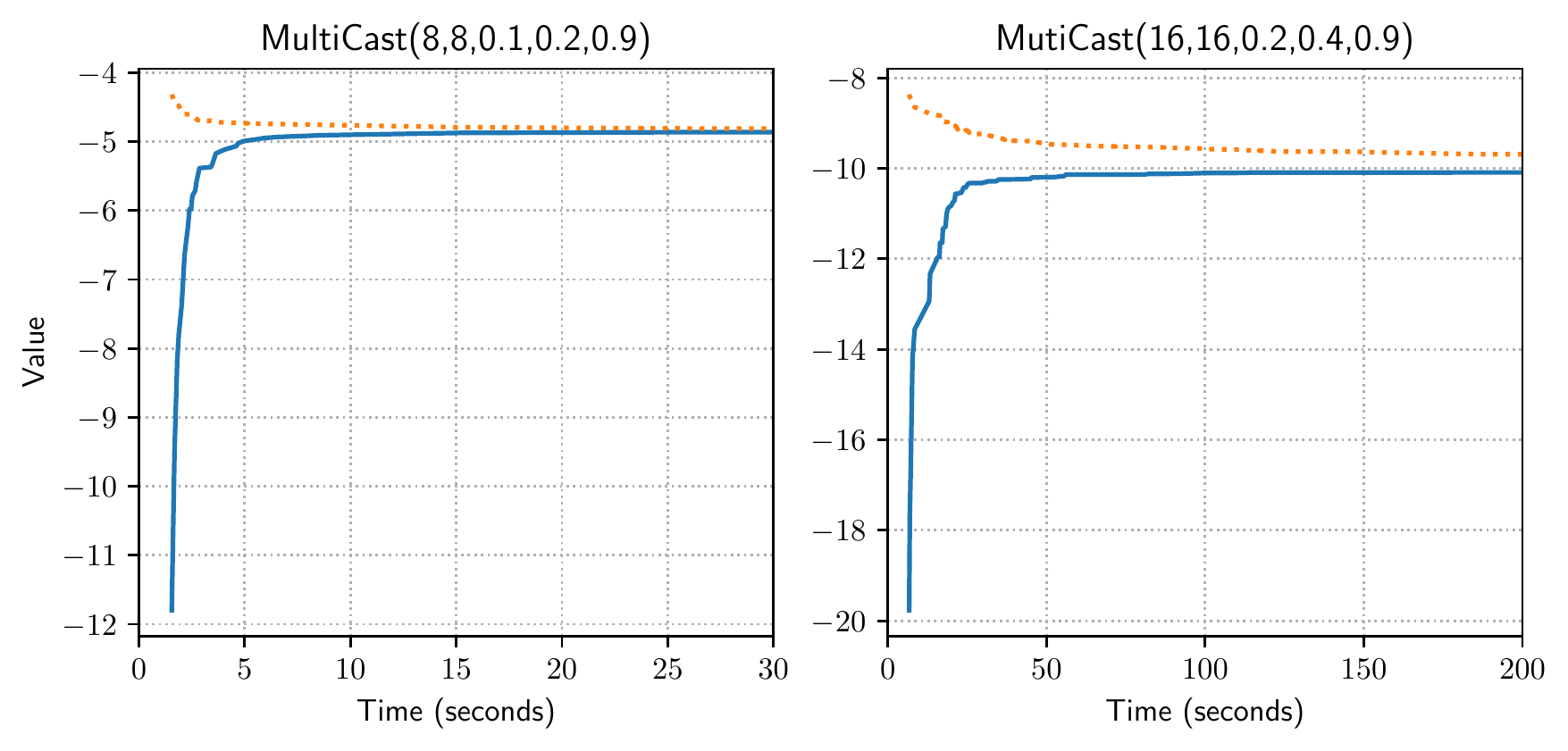}
    \caption{Experimental Results on \texttt{MultiCast}~($C^1, C^2, p^1, p^2, \beta$). The dotted line represents the upper bound $U(b_0)$ and the solid line represents the lower bound $L(b_0)$. Only the begining of the algorithm is shown. The lower bound quickly arrives at a near optimal value. For the rest of the run time, the upper bound and lower bound both moves very slowly.}
    \label{fig:multicast}
\end{figure}

\blue{Even though the algorithm took hours to close the gap between upper and lower bound to 0.01 for \texttt{MultiCast} instances, we would like to note that in hindsight, the lower bound arrives at a near optimal value very quickly as shown in Figure \ref{fig:multicast}. The remainder of the algorithm mostly reduces the upper bound to provide an optimality guarantee. Recall that CHSVI is an anytime algorithm. This means that even if we terminate the algorithm early, we can still obtain a strategy with near optimal performance.}

It took the algorithm more than 18 hours to reduce the gap for \texttt{DecTiger}(3,1,0.9). However, in hindsight, one can observe that the lower bound arrives at a near optimal value relatively quickly (At the 3 hour mark, the lower bound is already at 5.9919, not too far from the final value of 6.7236). In \texttt{DecTiger}(3,1,0.99), the gap is still large at the 24 hours mark. However, we conjecture that the solution reported at 24 hours is already close to optimal, and the gap will continue to decrease with run time.

\blue{We would like to note that directly applying off-the-shelf POMDP solvers to any of the instances we used is out of the question on our computer: We have tried to apply the state-of-the-art HSVI2 solver \cite{hsvi2} on \texttt{MultiCast}(5,5,0.1,0.2,0.9), which has $36$ states, $4$ observations, and $2^{12}$ actions/prescriptions. The program took more than 1 hour just to initialize the solver (in this stage, the algorithm does not output any solution if one terminates the algorithm).
In comparison, the instances we solved are much bigger: The number of states, observations, and actions/prescriptions for the coordinator's POMDP of our instances is listed in Table \ref{tab:dims}. Given that the complexity of the initialization stage of HSVI2 algorithm (which computes an upper bound with the Fast Informed Bound method \cite{hsvi2}) is linear in the number of actions/prescriptions, we project that it would take at least 2.7 days for \texttt{MultiCast}~(8,8,0.1,0.2,0.9) and 194 days for any other instances just to initialize the HSVI2 solver. 
\begin{table}[!ht]
    \centering
    \begin{tabular}{c|cccc}
         & $|\cS|$ & $|\cO|$ & $|\bcA|$  \\\hline
        \texttt{DecTiger}~(2,1,$\beta$)& 74 & 37 & $3^{14}$ \\
        \texttt{DecTiger}~(3,1,$\beta$)& 435 & 145 & $4^{26}$ \\
        \texttt{MultiCast}~(8,8,0.1,0.2,0.9)& 64 & 4 & $2^{18}$\\
        \texttt{MultiCast}~(16,16,0.2,0.4,0.9)& 256 & 4 & $2^{34}$
    \end{tabular}
    \caption{Dimensions of the coordinator's POMDP.}
    \label{tab:dims}
\end{table}
} 

\section{Conclusions}
\blue{
Decentralized stochastic control problems can be significantly more difficult than their centralized counterparts due to information asymmetry. Even though some decentralized control problem can be transformed into an equivalent centralized problem with the CI approach, these centralized problems are still difficult to solve due to the extremely large number of actions.} In this work, we proposed the Coordinator's Heuristic Search Value Iteration (CHSVI) algorithm, which combines the CI approach and point-based POMDP methods to solve multi-agent stochastic control problems. 
Our algorithm allows us to solve multi-agent control problems much more efficiently than directly using point-based algorithms for coordinator's POMDP (see Remarks 1 and 2). Further, our  approach suggests  several immediate future directions for further improving scalability such as:
(1) approximate upper bound update methods for CHSVI; (2) combination of the CI approach with other point-based algorithms such as SARSOP\cite{sarsop}; (3) efficient methods to initialize $\alpha$-constraints for the upper bound representation.

\bibliographystyle{IEEEtran}
\bibliography{ref}

\appendix
\subsection{Experimental Environments}
\blue{In this section, we introduce the environments we use for our experimental results in Section \ref{sec:experimental}.}

\subsubsection{DecTiger}\label{dectiger}
    We consider a parameterized extension of the DecTiger \cite{nair2003taming,dibangoye2016optimally} problem with three parameters $(N, d, \beta)$: 
    Two agents face $N$ closed doors. Behind one of the doors is a ferocious tiger, while behind all other doors are boxes of tiger-shaped cookies. At each time, each agent has $N+1$ choices: open one of the doors, or listen for the roar of the tiger. However, listening has a cost and is prone to error: an agent hears the roar from the correct door with probability $p = 0.85 / (0.7 + 0.15 N)$ and any other door with probability $q = 0.15 / (0.7 + 0.15 N)$. The instantaneous reward for the agents are listed in Table \ref{tab:dectiger}. The discount factor is $\beta$. Initially, the tiger is placed behind a uniform random door. The state of the world does not change if both agents choose to listen. If any door is opened by either agent, then the tiger resets itself to a uniform random location. If both agents choose to listen, then at the next time step, they will obtain conditionally independent observations each following the distribution described above. If either of the agents opens any door, then all observations will be independently uniform random and independent of the tiger's actual position, regardless of the other agent's action.

    \begin{table}[!ht]
        \centering
        \begin{tabular}{c|ccc}
             $a^1\backslash a^2$& listens & open tiger door & open other door \\\hline
             listens& $-2$ & $-101$ & $\frac{20}{N}-1$\\
             open tiger door& $-101$ & $-50$ & $-100$ \\
             open other door& $\frac{20}{N} - 1$ & $-100$ & $\frac{40}{N}$
        \end{tabular}
        \caption{Instantaneous rewards for \texttt{DecTiger}($N, d, \beta$)}
        \label{tab:dectiger}
    \end{table}
    
    In the original setup \cite{nair2003taming,dibangoye2016optimally}, the agents' actions and observations are private information, as it is the tradition for the Dec-POMDP model. In this work, we consider a delayed information sharing model of this problem: The agents share their actions and observations with each other but with a delay of $d > 0$. More specifically, at time $t$, each agent has access to the other agent's actions taken at anytime before (not including) $t-d$ and the observations yielded from these actions. For agent $i$, the actions taken after (including) time $t-d$ and their resulting observations are private at time $t$.

    \texttt{DecTiger}~($N, d, \beta$) can be modeled with our multi-agent control model introduced in Section \ref{sec:modelmulti} as follows: 
    $\mathcal{S} = [N]$ represents the $N$ possible locations of the tiger. $\mathcal{A}^i = [N+1]$ represent the choices of agent $i$ at any given time. Let $\mathcal{Z}^i=[N]$ represent the space of observations of agent $i$ and $\mathcal{Z} = \mathcal{Z}^1\times\mathcal{Z}^2$. Then $\mathcal{M}^i = \{\emptyset\}\cup \bigcup_{k=1}^d (\mathcal{Z}^i\times\mathcal{A}^i)^k$ is the space of actions and observations of agent $i$ that have not been known to the other agent yet. (The symbol $\emptyset$ represents the nonexistence of such action and observations at the beginning.) The common observation space $\mathcal{O} = \{\emptyset\}\cup (\mathcal{Z}\times \mathcal{A})$ is the space of newly shared actions and observations, which are the actions and observations $d$-steps ago. (The symbol $\emptyset$ represents that the first piece of information shared by the agents has not arrive yet.) The rest of the tuple making up the model $\mathcal{E}$ (i.e. $b_0, \Pr, r$) can be defined naturally.

\subsubsection{MultiCast}\label{multicast}
\blue{We consider a two-agent multi-access broadcast system model inspired by \cite{mahajan2013optimal} and \cite{pajarinen2011periodic}. The model is parameterized by $(C^1, C^2, p^1, p^2, \beta)$. Each agent $i$ has a buffer with size $C^i$. Initially, the buffers are empty. At each time $t$, a packet arrives at agent $i$'s buffer with probability $p^i$, independent of all history and the packet arrival event of the other agent. If the buffer is full, then the packet is dropped and the agents pay a penalty of $c_D = 2$. Otherwise, the packet is placed into the buffer. Both users may attempt to transmit a packet over a shared broadcast medium. If only one agent attempt to transmit, then the packet is successfully transmitted and removed from the buffer. If both agent attempt to transmit at the same time, then transmissions of both agents fail due to collision. An attempt to transmit, whether successful or not, costs $c_T = 0.5$. In addition to the drop penalty and transmission cost, the agents pay a penalty of 1 for each unit of time each packet stays in the buffer. The discount factor for the penalties and costs is $\beta$. The agents know their own packet arrival histories but not the other agents. An agent can also observe the other agent's action after it is taken. 

It can be shown that \cite{mahajan2013optimal,hamidsuffinfo} in this model, the number of packets in the buffer for each agent $i$ is a \emph{sufficient private information} for decision making. Therefore, per Remark \ref{remark:suffinfo}, \texttt{MultiCast}~($C^1, C^2, p^1, p^2, \beta$) can be modeled with our multi-agent control model in Section \ref{sec:modelmulti} as follows: $\mathcal{S}^i = \{0, 1, \cdots, C^i\}$ is the set of private states for agent $i$ indicating the number of packets in its buffer. $\mathcal{S} = \mathcal{S}^1\times\mathcal{S}^2$ is the state space. $\mathcal{A}^i = \{\mathrm{NT}, \mathrm{T}\}$ represent the two choices of agent $i$ at any given time. $\cM^i = \cS^i$ is the space of sufficient private information. $\cO = \cA$ is the space of common observations. The instantaneous reward $r$ is defined by
\begin{align}
    r(s, a) &= \sum_{i=1}^2 r^i(s^i, a^i)\\
    r^i(s^i, a^i) &= -s^i -c_D\cdot p^i \bm{1}_{\{ s^i = C^i \}} -c_T\cdot\bm{1}_{\{a^i=\mathrm{T}\}}
\end{align}
The rest of the tuple making up the model $\mathcal{E}$ (i.e. $b_0, \Pr$) can be defined naturally.
}

\subsection{More on the New Upper Bound Representation}\label{app:morealphaconstr}
In this section we describe more types of $\alpha$-constraints for the upper bound representation described in Section \ref{sec:ubrepr}.

\subsubsection{Lower Bound Based Constraints} For any belief $b$, we can derive a \emph{lower bound} $\underline{v}^b$ on $\min_{\alpha\in \cV^*}\alpha^T b$ through considering a reward \emph{minimization} POMDP. Then, we can add the inequality $b^T y \geq \underline{v}^b$ to the set of $\alpha$-constraints. A crude lower bound of this kind can be efficiently computed through the Fast Informed Bound method \cite{hauskrecht2000value}.

\subsubsection{Marginal Belief Based Constraints}\label{app:mbbalpha} Oftentimes, we can derive an upper bound of $V^*(b)$ as a function of some marginal distribution of $b$, i.e. suppose that $\cS = \mathcal{K}^1\times \mathcal{K}^2$, we have $V^*(b) \leq \Bar{v}(\varphi(b))$ for all $b\in\Delta(\cS)$ where $\varphi(b)\in\Delta(\mathcal{K}^1), [\varphi(b)](k^1) = \sum_{k^2\in\mathcal{K}^2} b(k^1, k^2)$. An example is the upper bound for a coordinator's POMDP derived by solving a relaxed problem where part or all of the private information is assumed to be commonly known. (The relaxed problem can be much easier to solve due to the fact that it features a smaller private information space and a smaller augmented state space.) This kind of upper bound can be easily translated into $\alpha$-constraints: Suppose that we have an upper bound $\Bar{v}^{b^1}$ at a fixed $b^1\in \Delta(\mathcal{K}^1)$, then we know that all $\alpha$-vectors in $\cV^*$ satisfy the constraint
\begin{equation}\label{eq:maxbty}
    \max_{b:\varphi(b)=b^1} b^T y \leq \Bar{v}^{b^1}.
\end{equation}

We can transform \eqref{eq:maxbty} into a group of finitely many linear constraints through the following steps: First note that 
$$\max_{b:\varphi(b)=b^1} b^T y = \sum_{k^1\in\mathcal{K}^1} b^1(k^1) \max_{k^2\in\mathcal{K}^2} y(k^1, k^2).$$
Therefore, by adding an auxiliary vector $\Bar{y}\in \mathbb{R}^{\mathcal{K}^1}$, we obtain the $\alpha$-constraints:
\begin{align}
    y(k^1, k^2)&\leq \Bar{y}(k^1)\quad\forall k^1\in\mathcal{K}^1, k^2\in\mathcal{K}^2,\label{eq:yleqz}\\
    (b^1)^T \Bar{y} &\leq \Bar{v}^{b^1}.
\end{align}

In the case that a lower bound for the reward minimization problem can be expressed as a function of the marginal belief, one can use a similar way to obtain the corresponding $\alpha$-constraints as well.
\blue{
Suppose that $\underline{v}^{b^1}$ is a lower bound of the minimum total reward at belief $b$ for all $b$ such that $\varphi(b) = b^1$, then we can use an auxiliary vector $\underline{y}\in \mathbb{R}^{\mathcal{K}^1}$ to formulate the $\alpha$-constraints:
\begin{align}
    y(k^1, k^2)&\geq \underline{y}(k^1)\quad\forall k^1\in\mathcal{K}^1, k^2\in\mathcal{K}^2,\label{eq:ygeqz}\\
    (b^1)^T \underline{y} &\geq \underline{v}^{b^1}.
\end{align}
} 

\subsection{Additional Figures for Experimental Results}\label{app:experiment}
\begin{figure}[!ht]
    \centering
    \includegraphics[width=0.49\textwidth]{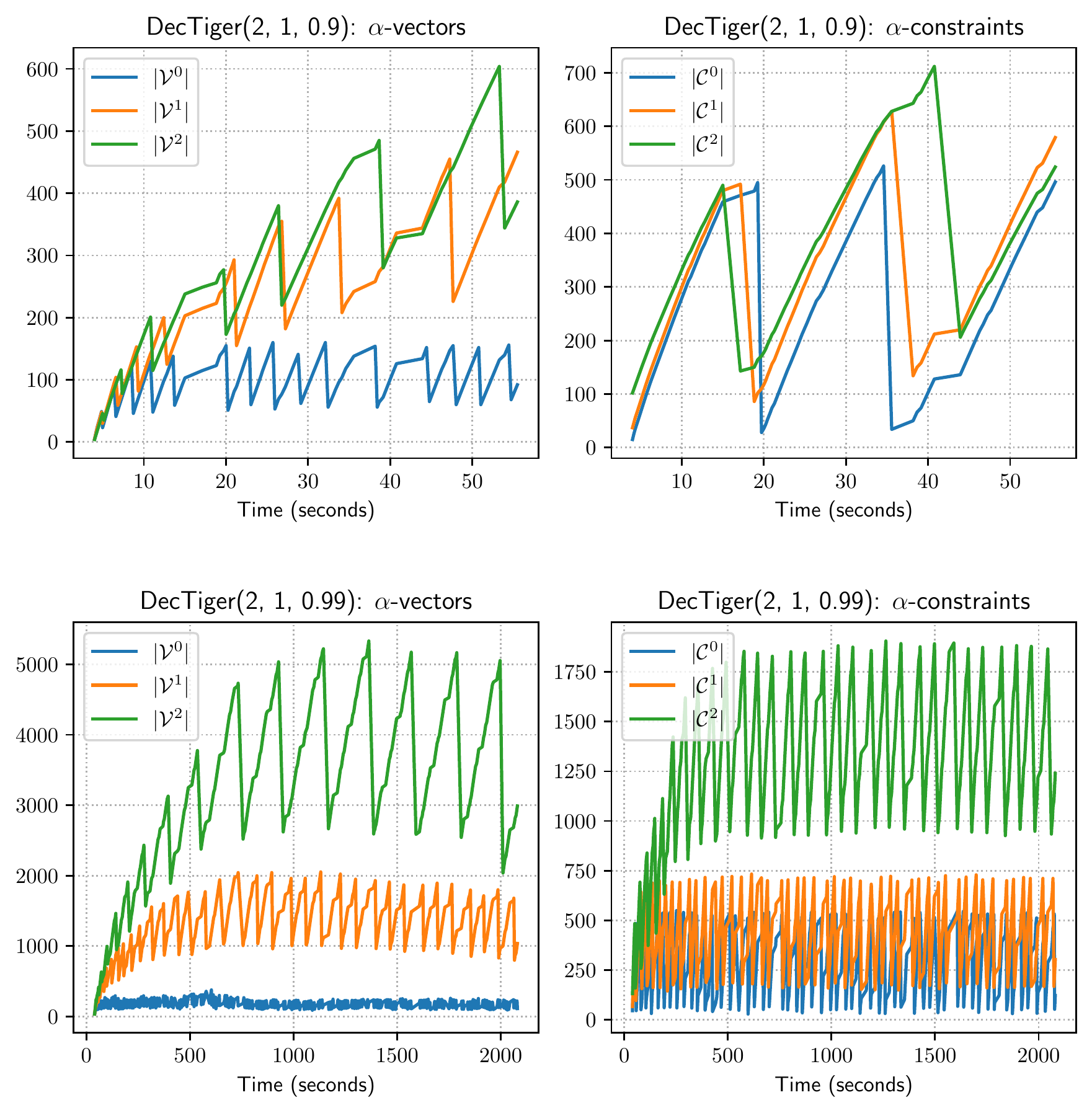}
    \caption{Number of $\alpha$-vectors and $\alpha$-constraints for \texttt{DecTiger}~($2, 1, \beta$). $\cV^\ell$ is the set of $\alpha$-vector for stage $\ell$ defined in Algorithm \ref{alg:chsvi:lbupdate}. $\mathcal{C}^\ell$ is the set of $\alpha$-constraints for stage $\ell$ defined in Algorithm \ref{alg:chsvi:ubupdate}.}
    \label{fig:dectigerstats2}
\end{figure}

\begin{figure}[!ht]
    \centering
    \includegraphics[width=0.49\textwidth]{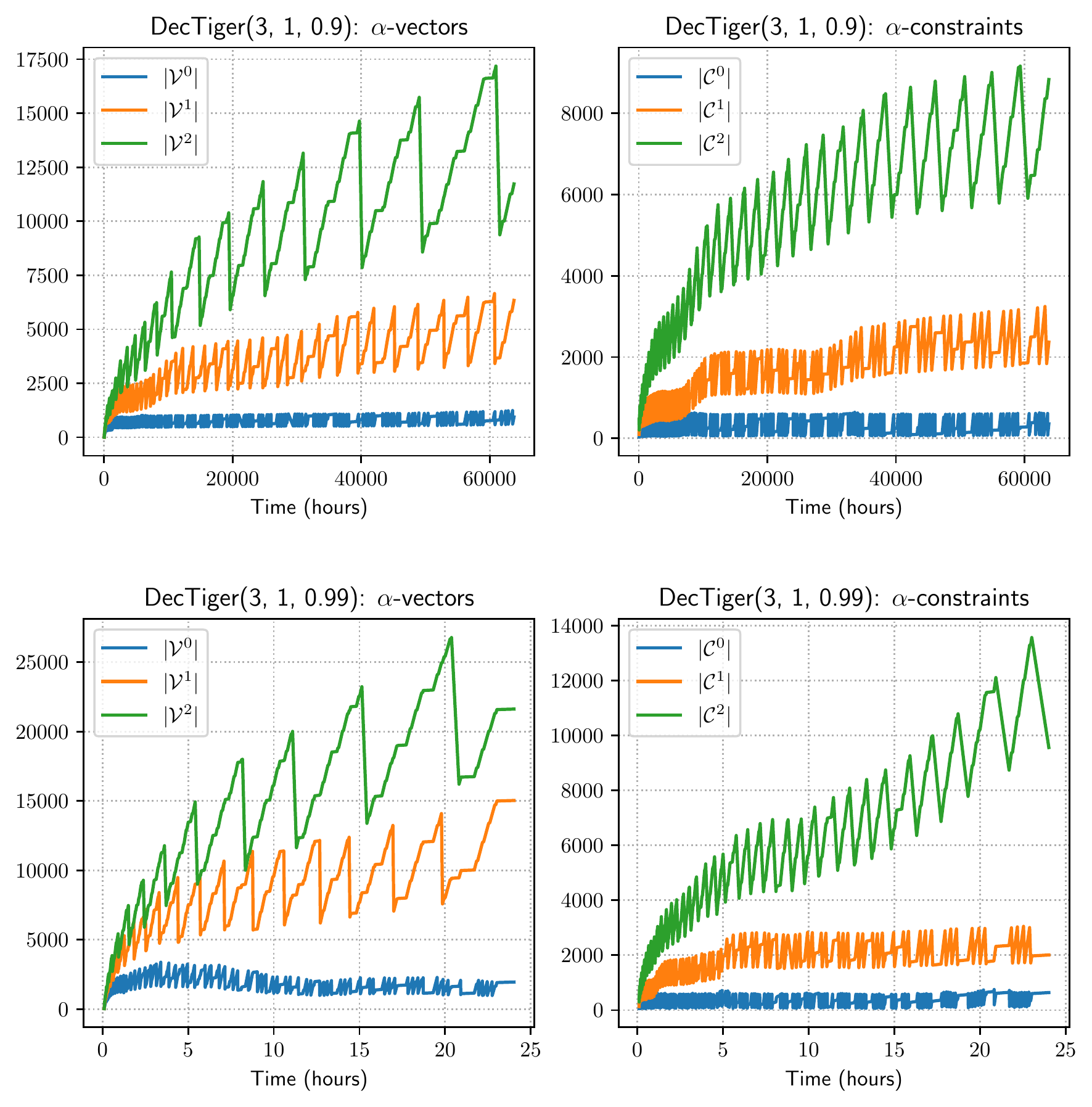}
    \caption{Number of $\alpha$-vectors and $\alpha$-constraints for \texttt{DecTiger}~($3, 1, \beta$). $\cV^\ell$ is the set of $\alpha$-vector for stage $\ell$ defined in Algorithm \ref{alg:chsvi:lbupdate}. $\mathcal{C}^\ell$ is the set of $\alpha$-constraints for stage $\ell$ defined in Algorithm \ref{alg:chsvi:ubupdate}.}
    \label{fig:dectigerstats3}
\end{figure}

\blue{In the CHSVI algorithm, the memory use is mainly from two sources: (1) $\alpha$-vectors for the lower bound representation, and (2) $\alpha$-constraints for the upper bound representation. We record the number of $\alpha$-vectors and $\alpha$-constraints\footnote{$\alpha$-constraints that relate auxiliary variables to main variables, i.e. \eqref{eq:yleqz}\eqref{eq:ygeqz}, are not included.} during each run of the algorithm. The results are shown in Figure \ref{fig:dectigerstats2} and \ref{fig:dectigerstats3}. The zigzag pattern is due to periodic pruning of $\alpha$-vectors and $\alpha$-constraints.

As is the case for HSVI on single-agent POMDPs, the pruning procedure plays a crucial role in reducing the and memory complexity of the algorithm. It also play a significant role in reducing the time complexity since stage optimization problems with fewer $\alpha$-vectors or fewer $\alpha$-constraints are easier to solve. From the figures, we observe that in all cases, the majority of $\alpha$-vectors and $\alpha$-constraints are those for stage $2$. This is to be expected since stage $2$ has the largest state space of the three stages, resulting in a large dimension of the belief space. Since a larger state space also means larger memory use for each $\alpha$-vector or $\alpha$-constraint, the stage 2 $\alpha$-vectors and $\alpha$-constraints dominate the memory use. In \texttt{DecTiger}(2,1,0.99) and \texttt{DecTiger}(3,1,0.9), we observe that the number of $\alpha$-vectors and $\alpha$-constraints grows with a decreasing rate until it stabilizes well before the termination of the algorithm. We conjecture that the same behaviour holds for \texttt{DecTiger}(3,1,0.99) if it were given enough run time.
}

\end{document}